\documentclass[10pt,letterpaper]{article}

\usepackage{cogsci}

\cogscifinalcopy % Uncomment this line for the final submission 

\usepackage{pslatex}
\usepackage{apacite}
\usepackage{float} % Roger Levy added this and changed figure/table
\usepackage{graphicx}
\usepackage{subcaption}
\usepackage{amssymb}
\usepackage{booktabs}
\usepackage{multirow}
\usepackage{tabularx}
\usepackage{fancyhdr}
\title{Simulating Problem Difficulty in Arithmetic Cognition Through \\ Dynamic Connectionist Models}
\author{
    {\large \bf Sungjae Cho\textsuperscript{\normalfont 1} (sj.cho@snu.ac.kr), Jaeseo Lim\textsuperscript{\normalfont 1} (jaeseolim@snu.ac.kr),} \\
    {\large \bf Chris Hickey\textsuperscript{\normalfont 1} (chris.hickey@ucdconnect.ie), Jung Ae Park\textsuperscript{\normalfont 2} (lydia120@snu.ac.kr),} \\
    {\large \bf Byoung-Tak Zhang\textsuperscript{\normalfont 1,3} (btzhang@bi.snu.ac.kr)} \\
    \textsuperscript{1} Interdisciplinary Program in Cognitive Science, Seoul National University, \\
    \textsuperscript{2} Department of Psychology, Seoul National University, \\
    \textsuperscript{3} Department of Computer Science and Engineering, Seoul National University, \\
    1, Gwanak-ro, Gwanak-gu, Seoul, 08826, Republic of Korea
}

\begin{document}

% comment below for submission version ===========================
\fancyhf{} % clear all header and footers
\renewcommand{\headrulewidth}{0pt} % remove the header rule
\pagestyle{fancy}
\cfoot{\vspace*{1.5\baselineskip}\thepage}
% =================================================================
\maketitle

\begin{abstract}
The present study aims to investigate similarities between how humans and connectionist models experience difficulty in arithmetic problems.
Problem difficulty was operationalized by the number of carries involved in solving a given problem. 
Problem difficulty was measured in humans by response time, and in models by computational steps.
The present study found that both humans and connectionist models experience difficulty similarly when solving binary addition and subtraction.
Specifically, both agents found difficulty to be strictly increasing with respect to the number of carries.
Furthermore, the models mimicked the increasing standard deviation of response time seen in humans.
Another notable similarity is that problem difficulty increases more steeply in subtraction than in addition, for both humans and connectionist models.
Further investigation on two model hyperparameters --- confidence threshold and hidden dimension --- shows higher confidence thresholds cause the model to take more computational steps to arrive at the correct answer. 
Likewise, larger hidden dimensions cause the model to take more computational steps to correctly answer arithmetic problems; however, this effect by hidden dimensions is negligible.

\end{abstract}

\section{Introduction}
%Cognitive arithmetic studies the mental representation of numbers and arithmetic, and the processes that access and use this knowledge \cite{Ashcraft1992_ProblemDifficulty}.

%[Mathematical Cognition] 
%[Problem Difficulty and Carries]
%% 1
%Mathematical cognition is the field of research concerned with the cognitive processes that underlie mathematical abilities \cite{Campbell2005_Handbook}. 
%The field involves a subfield named cognitive arithmetic that studies the mental representation of numbers and arithmetic, and the processes that access and use this knowledge \cite{Ashcraft1992_ProblemDifficulty}.
%\textit{Problem difficulty} is a central variable in cognitive arithmetic \cite{Ashcraft1992_ProblemDifficulty, Ashcraft1995_ProblemDifficulty}. 
%% 2
Do connectionist models experience difficulty on arithmetic problems like humans?
Although connectionist models consist of abstract biological neurons, similar behaviors between humans and these models are not guaranteed. % due to their high abstraction.
However, developing model simulations to discover such similarities can bridge this knowledge gap between humans and models, and deepen our understanding of the micro-structures involved in cognition \cite{RumelhartM86,Mcclelland1988_connectionist}.
Therefore, finding such similarities is a foundational step in understanding human cognition through connectionist models.
%This connectionist approach recently has been used in the domain of mathematical cognition \cite{McclellandMHYL16, Saxton2019_mathDataset}.
This connectionist approach recently has been used in the domain of mathematical cognition \cite{McclellandMHYL16, Mickey2014, Saxton2019_mathDataset}.

%Then, what problems humans feel difficult?
Cognitive arithmetic \cite{Ashcraft1992_ProblemDifficulty}, the study of the mental representation of arithmetic, conceptualizes \textit{problem difficulty}.
Problem difficulty can be measured by \textit{response time} (RT) from the time a participant sees an arithmetic problem to the time the participant answers the problem \cite{Imbo2007_NumCarry}.
%--- how difficult arithmetic problems are.
%Problem difficulty is a central variable in cognitive arithmetic \cite{Ashcraft1992_ProblemDifficulty, Ashcraft1995_ProblemDifficulty}.
%There are at least three causes for operationalizing problem difficulty: 

There are three criteria that affect problem difficulty \cite{Ashcraft1992_ProblemDifficulty, Imbo2007_NumCarry}:
(a) operand magnitude (e.g., 1 + 1 vs. 8 + 8); (b) number of digits in the operands (e.g., 3 + 7 vs. 34 + 78); and (c) the number of carry\footnote{A \textit{carry} in binary addition is the leading digit $1$ shifted from one column to a more significant column when the sum of the less significant column exceeds a single digit. A \textit{borrow} in binary subtraction is the digit 1  shifted to a less significant column in order to obtain a positive difference in that column. We refer to borrows as carries.} operations (e.g., 15 + 31 vs. 19 + 37). 
%There are at least three factors that effect problem difficulty: 
%(a) operand magnitude (e.g., 1 + 1 vs. 8 + 8); (b) number of digits in the operands (e.g., 3 + 7 vs. 34 + 78); and (c) the presence or absence of carries\footnote{A \textit{carry} in binary addition is the leading digit $1$ shifted from one column to a more significant column when the sum of the less significant column exceeds a single digit. A \textit{borrow} in binary subtraction is the digit 1  shifted to a less significant column in order to obtain a positive difference in that column. This paper refers to borrows as carries.} operations (e.g., 15 + 31 vs. 19 + 37). 
%In particular, factor (c) has been investigated with regard to the number of carries in a problem \cite{FurstH00, Imbo2007_NumCarry}. 
%[X]In the present study, we investigate how the number of carries affects problem difficulty. 
%As most preceding studies measure problem difficulty through \textit{response time} (RT) from the time a participant sees a problem to the time the participant answer to the problem, the present study adopts this measure.
%[Binary numeral system]
%Previous studies that examine the ways humans process numbers are mostly based on the highly familiar decimal numeral system. 
% The present study controls for the effects of criteria (a) and (b) on problem difficulty by using the experimental approach created by \citeA{Cho2019}. This approach adopts a binary numeral system, which allows for 
The present study uses a similar experimental approach to that suggested by \citeA{Cho2019}. This design employs the binary numeral system to control for familiarity with the decimal system and the two criteria (a) and (b). As such, the present study considers the number of carries as the only independent variable involved in problem difficulty.% An additional benefit to using binary numerals is that they are well suited for vectorization as input into connectionist models.
%In the present study, the number of carries is considered as the only dependent variable of problem difficulty.
%Morever, the binary numeral system is adopted to observe better the effect of carries on the human's problem difficulty \cite{Cho2019}, and to help us less care about the representation of digits for connectionist models.
%The use of the binary system offers a novel way to ensure that any effect from previous experience with conventional mathematical operations are mitigated. 
%Moreover, because the binary system uses only 0 or 1 digits, it reduces the \textit{problem size effect}; criterion (a): problems with smaller operands (e.g., $5+2$, $4-1$) are solved more quickly and accurately than problems with larger operands (e.g., $7+6$, $9-6$) \cite{Campbell94_ProblemSizeEffect, LeFevre96_ProblemSizeEffect, Miller84_ProblemSizeEffect}. 
%Therefore, to observe the effect of carries on problem difficulty, the present study employs the binary system to control for familiarity with the decimal number system and the two causes (a) and (b).  

Recurrent neural networks \cite{Elman1990_ElmanNetwork, Jordan1997_JordanNetwork} can model sequential decisions through time.
These networks perform sequential nonlinear computations. 
Owing to the principle that many nonlinear computational steps are required to learn complex mappings \cite{Lecun2015_deepLearning}, parallels can be drawn between human RT and model computational steps in response to problems of varying difficulty level.

%\begin{figure}[tb!]
%\begin{center}
%\includegraphics[width=0.49\textwidth]{fig/modeling_diagram.pdf}
%\caption{Experimental phase diagram} 
%\label{fig/modeling_diagram}
%\end{center}
%\end{figure}

%Two ex[er]
%To answer our research question, experiments are conducted: one on human participants and the other on connectionist models.
Two experiments were conducted in the present study: one on human participants and the other on connectionist models.
Both experiments had \textit{learning} and \textit{solving} phases.
In the learning phase of the human experiment, participants were taught a method for solving binary arithmetic problems by following guiding examples.
In the solving phase, participants began the experiment in earnest, solving arithmetic problems under experimental conditions and having their RTs recorded as a measure of problem difficulty.
In the learning phase of the model experiment, connectionist models were trained until they achieved 100\% accuracy across all problems. We consider this to be roughly equivalent to how participants were taught to solve arithmetic problems in the learning phase of the human experiment.
In the solving phase, all problems were solved again and the number of computational steps taken to solve each problem were recorded as a measure of problem difficulty.
%The models are trained until they attains 100\% accuracy for all arithmetic problems.
%That is because we consider the state can be compared with the state humans have learned the algorithmic method that enables them to solve correctly all problems.
Following both experiments, results were analyzed in order to investigate whether any similarities could be observed in how both agents underwent problem difficulty with respect to the number of carries. 
We then investigated how major model configurations affect model behavior.

\section{Problem Sets}

\subsubsection{Operation datasets}

For addition and subtraction, we constructed separate \textit{operation datasets}, containing all possible operations between two 4-digit binary nonnegative integers that generate nonnegative results.
The addition dataset has 256 operations, and the subtraction dataset has 136 operations (Figure \ref{fig/problem_sets}).
%\footnote{Let us simply refer to the operation dataset of addition as the addition dataset, and that of subtraction as the subtraction datatset.} 
%\footnote{and problems from the subtraction dataset as subtraction problems.}
Operation datasets consist of $(\textbf{x},\textbf{y})$ where $\textbf{x}$ is an 8-dimensional input vector that is a concatenation of two binary operands, and $\textbf{y}$ is an output vector that is the result of computing these operands.
$\textbf{y}$ is 5-dimensional for addition and 4-dimensional for subtraction.

\subsubsection{Carry datasets}

\begin{figure}[tb!]
\begin{center}
\includegraphics[width=0.48\textwidth]{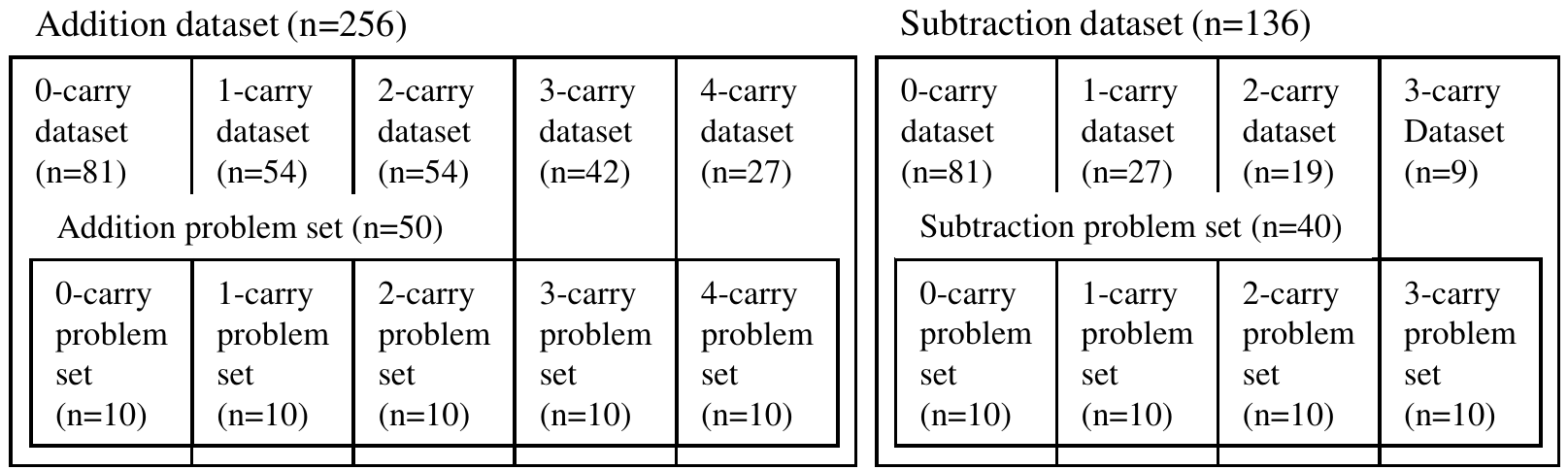}
\caption{Problem sets. The addition and subtraction datasets were assigned to connectionist models. The addition and subtraction problem sets were assigned to participants. $n$ refers to the number of operations in a given dataset/problem set.} 
\label{fig/problem_sets}
\end{center}
\end{figure}

% Need to explain more about carry datasets and the number of carries
% This part is not very organized.
Operation datasets were further subdivided into carry datasets.
A \textit{carry dataset} refers to the total set of operations in which a specific number of carries is required for a given operator. 
The addition dataset was divided into 5 carry datasets, and the subtraction dataset was divided into 4 carry datasets (Figure \ref{fig/problem_sets}).
%With $n_{c}$ denoting the number of carries required to correctly solve a problem, subtraction has 4 possible $n_{c}$. 
%Hence, subtraction has 4 carry datatsets. 
%The number of carry datasets for the other operators are shown in Table \ref{table/problem_sets}.
%Let us denote the number of possible $n_{c}$ as $N$. Depending on $N$, each operation dataset has $N$ carry datasets. 
For example, in Figure \ref{fig/guiding_examples}, the addition guiding examples (a) and (b) are in 2-carry\footnote{Let us simply refer to the carry dataset involving $n$ carries as the $n$-carry dataset, and problems from the $n$-carry dataset as $n$-carry problems.} and 4-carry datasets, respectively; 
the subtraction guiding examples (c) and (d) are in 2-carry and 3-carry datasets, respectively.  

\section{Experiment 1: Humans}

Experiment 1 investigated whether human RT in problem solving increases as a function of the number of carries involved in a problem.
%Experiment 1 aims to investigate whether humans take longer response time to solve a problem as the problem involves more carries.

\subsection{Participants}

90 undergraduate and graduate students (48 men, 42 women) from various departments completed the experiment. The average age of participants was 23.6 ($SD=3.3$).
% and the average self-reported math ability was 3.0 ($SD=0.6$) out of 4 (1: very bad; 2: below average; 3: above average; 4: very good).

\subsection{Materials}

Participants were given two types of problem sets: addition and subtraction. 
The addition problem set was constructed as follows: 
10 different problems were sampled from each carry dataset without replacement\footnote{
    This only occurred when sampling 3-carry problems ($n=10$) from the 3-carry subtraction dataset ($n=9$). This required one random problem to be duplicated and shown twice in the 3-carry problem set.
}.
These sampled problems were shuffled together to make the addition problem set.
%Then, the shuffled problems became the addition problem set,
This addition problem set was comprised of 50 unique problems evenly distributed across 5 carry datasets (Figure \ref{fig/problem_sets}).
%Then, the shuffled problems became the problem set, which has 50 unique problems because there are 5 carry datasets (Figure \ref{fig/problem_sets}).
Likewise, the subtraction problem sets consisted of 40 problems evenly distributed across 4 carry datasets (Figure \ref{fig/problem_sets}).
%In the case of subtraction, the problem set has 40 problems (Figure \ref{fig/problem_sets}).
The problems were newly sampled for each participant.

In any given problem, two operands were presented in a fixed 4-digit format in order to control for possible extraneous influences on problem difficulty %\cite{Ashcraft1992_ProblemDifficulty, Ashcraft1995_ProblemDifficulty}, 
as outlined by criterion (b).
The experiment was designed in such a way that participants were required to fill out all digits when answering questions (e.g. if the answer was 1, participants were forced to respond with 0001 as opposed to just 1).
This is to ensure RT is not affected by the number of answer digits.
%In solving a problem, participants must click either 0 or 1 for all 5 digits (addition) or 4 digits (subtraction) of their answer, even though its leading digits are zero, and then click the submit button.
% Paragraph in CogSci2019 %%
%Question distributions per problem set were as follows: 
%addition - 50 problems across 5 carry datasets; 
%subtraction - 40 problems across 4 carry datasets. 
%Problems were evenly distributed across carry problem sets, such that participants answered equal numbers of questions from the 0-carry problem set, 1-carry problem set, and so on .
%%%%

\begin{figure}[tb!]
    \centering
    \begin{subfigure}[b]{0.12\textwidth}
        \centering
        \includegraphics[width=\textwidth]{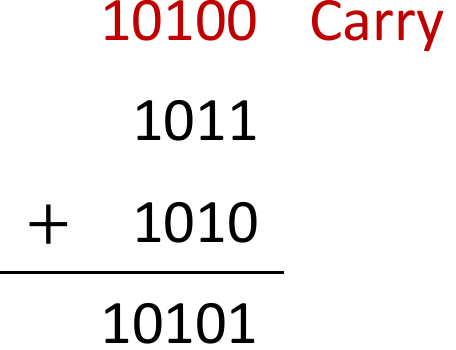}
        \caption{}
        \label{fig/guiding_example_1}
    \end{subfigure}%\hfill
    \begin{subfigure}[b]{0.12\textwidth}
        \centering
        \includegraphics[width=\textwidth]{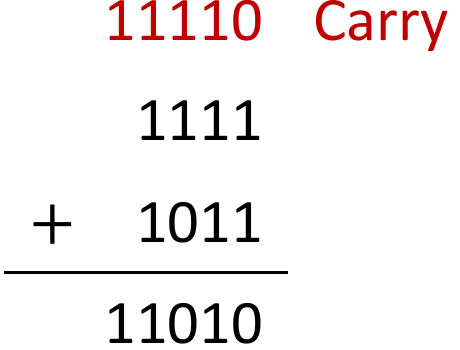}
        \caption{}
        \label{fig/guiding_example_2}
    \end{subfigure}%\hfill
    \begin{subfigure}[b]{0.12\textwidth}
        \centering
        \includegraphics[width=\textwidth]{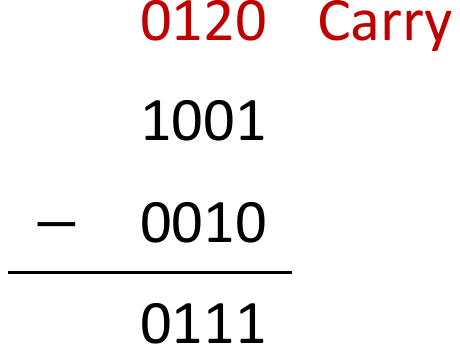}
        \caption{}
        \label{fig/guiding_example_3}
    \end{subfigure}%\hfill
    \begin{subfigure}[b]{0.12\textwidth}
        \centering
        \includegraphics[width=\textwidth]{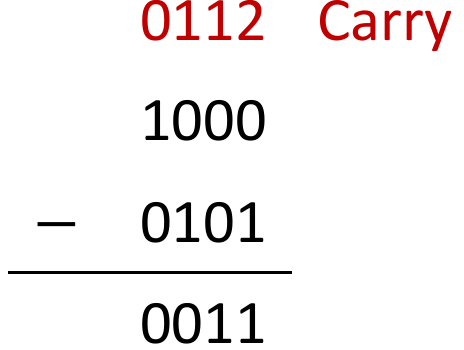}
        \caption{}
        \label{fig/guiding_example_4}
    \end{subfigure}%\hfill
    \caption{Guiding examples}
    \label{fig/response_time}
    \label{fig/guiding_examples}
\end{figure}

\subsection{Procedure}

Participants were shown calculation guidelines containing two guiding examples for addition (Figure \ref{fig/guiding_example_1}, \ref{fig/guiding_example_2}).
%, and instructions to understand the program interface. 
Participants were explicitly requested to solve problems by using carry operations outlined in the examples.
Participants then began to solve each problem from their addition problem set. 
After solving all addition problems, participants repeated the previous procedure for their subtraction problem set with two subtraction guiding examples (Figure \ref{fig/guiding_example_3}, \ref{fig/guiding_example_4}). 
%After solving all addition problems, participants learned the same guidelines including two guiding examples for subtraction (Figure \ref{fig/guiding_example_3}, \ref{fig/guiding_example_4}). 
%Then, they started to solve each problem from a subtraction problem set. 
Participants were prohibited from using any writing apparatus in order to force participants to solve problems mentally. 

\subsection{Results}

Analysis of variance (ANOVA) was used to investigate differences in mean RTs of participants across carry problem sets.
If there were significant differences between all the mean RTs, post hoc analysis was applied.
If a participant provided a wrong answer, it was reasonable to assume that this participant made some cognitive error when solving the problem.
As such, only RTs for correct answers were included in analysis.
We removed the outlying RTs of each carry problem set for each participant since unusually short RTs may be due to memory retrieval and excessively long RTs may be caused by distraction or anxiety during problem solving.
The RTs in the range $[Q_{1}-1.5\cdot\textup{IQR}, Q_{3}+1.5\cdot\textup{IQR}]$ were considered outliers, where $Q_{1}$ and $Q_{3}$ were the first and third quantiles of the RTs for a carry problem set, and $\textup{IQR}=Q_{3}-Q_{1}$.
%From here on, let us simply refer to the mean RT of a participant as RT (Change!!).  

\subsubsection{Addition}

\begin{figure}[bt!]
    \centering
    \begin{subfigure}[b]{0.25\textwidth}
        \centering
        \includegraphics[height=4.0cm]{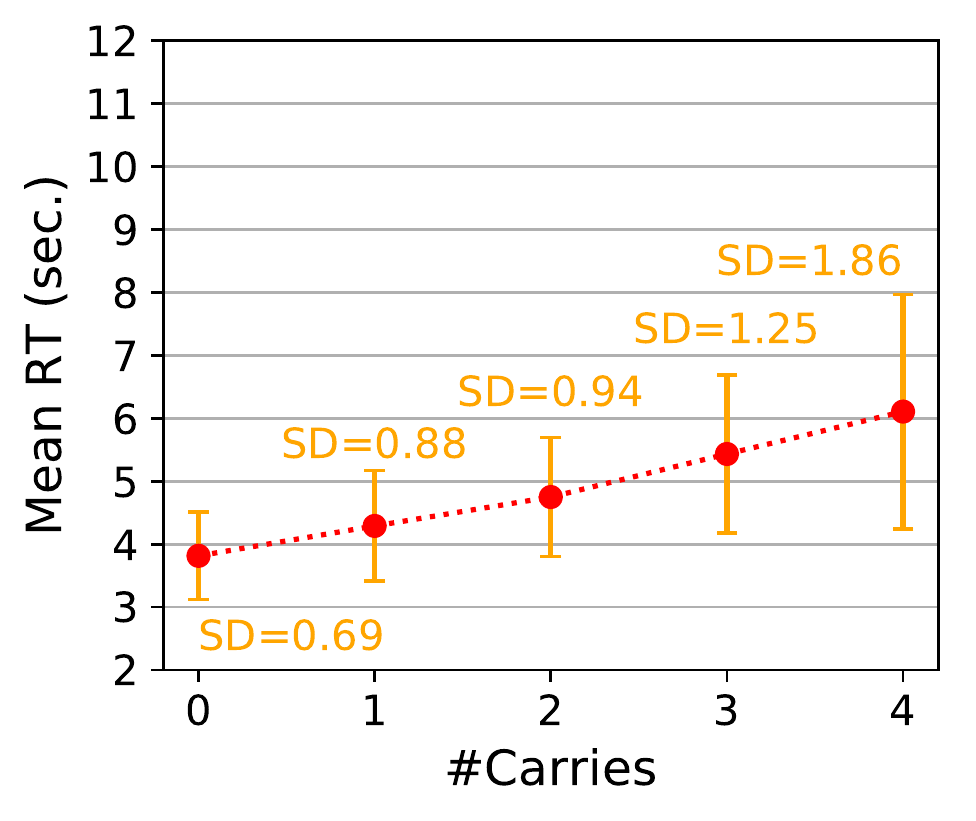}
        \caption{Addition}
        \label{fig/response_time_by_carries_add}
    \end{subfigure}%\hfill
    \begin{subfigure}[b]{0.25\textwidth}
        \centering
        \includegraphics[height=4.0cm]{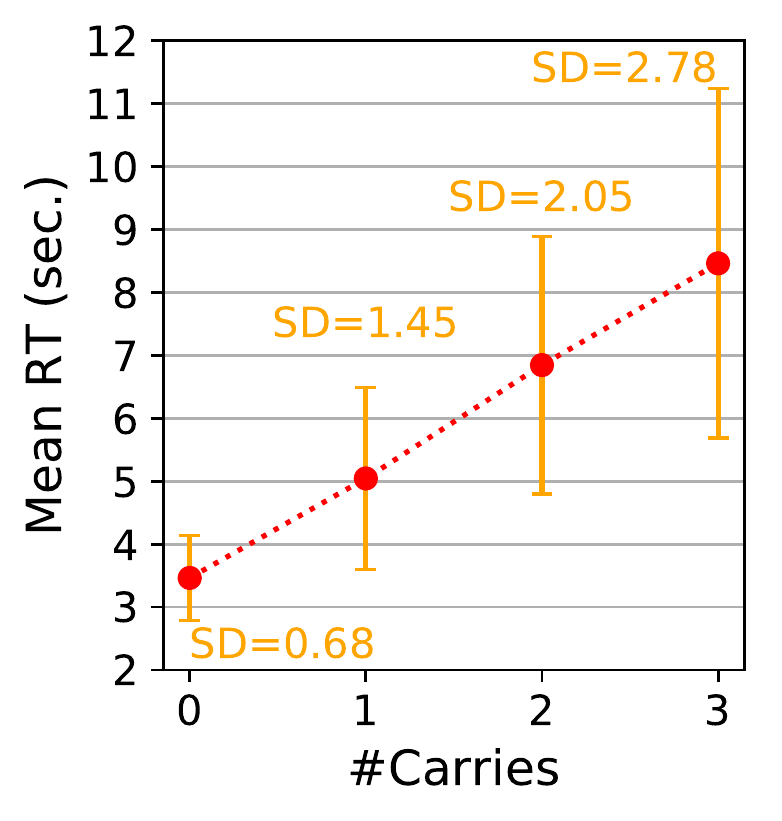}
        \caption{Subtraction}
        \label{fig/response_time_by_carries_subtract}
    \end{subfigure}%\hfill
    \caption{Mean RT by carries. The error bars are $\pm 1SD$.}
    \label{fig/response_time_by_carries}
\end{figure}

There were significant differences in mean RTs between all carry problem sets, as determined by ANOVA [$F(4, 445)=51.84, p<.001$, $\eta^{2}=.32$].
Post hoc comparisons using the Games-Howell test indicated that mean RTs between any two carry problem sets showed a significant difference [3-carry and 4-carry problem sets: $p=.040$; other pairs: $p<.01$].
Therefore, the mean RT was strictly increasing
\footnote{For every $x$ and $x'$ such that $x<x'$, if $f(x) < f(x')$, then we say $f$ is \textit{strictly increasing}.} with respect to the number of carries (Figure \ref{fig/response_time_by_carries_add}).

\subsubsection{Subtraction}

There were significant differences in mean RTs between all carry problem sets, as determined by ANOVA [$F(3, 356)=117.41$, $\eta^{2}=.50$].
Post hoc comparisons using the Games-Howell test indicated that mean RTs between any two carry problem sets showed a significant difference [$p<.001$].
Therefore, the mean RT was strictly increasing with respect to the number of carries (Figure \ref{fig/response_time_by_carries_subtract}).
\begin{figure*}[ht!]
    \centering
    \begin{subfigure}[b]{0.25\textwidth}
        \centering
        \includegraphics[height=5.5cm]{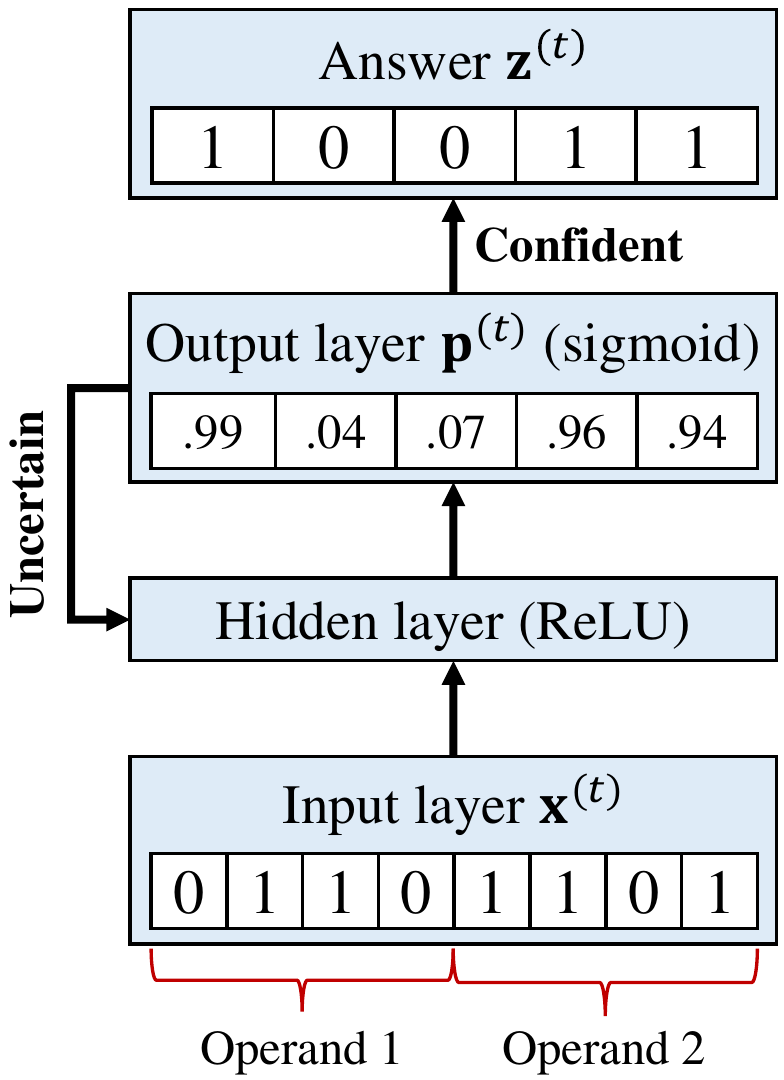}
        \caption{The Jordan network for addition}
        \label{fig/jordan_network}
    \end{subfigure}%\hfill
    \begin{subfigure}[b]{0.75\textwidth}
        \centering
        \includegraphics[height=5.7cm]{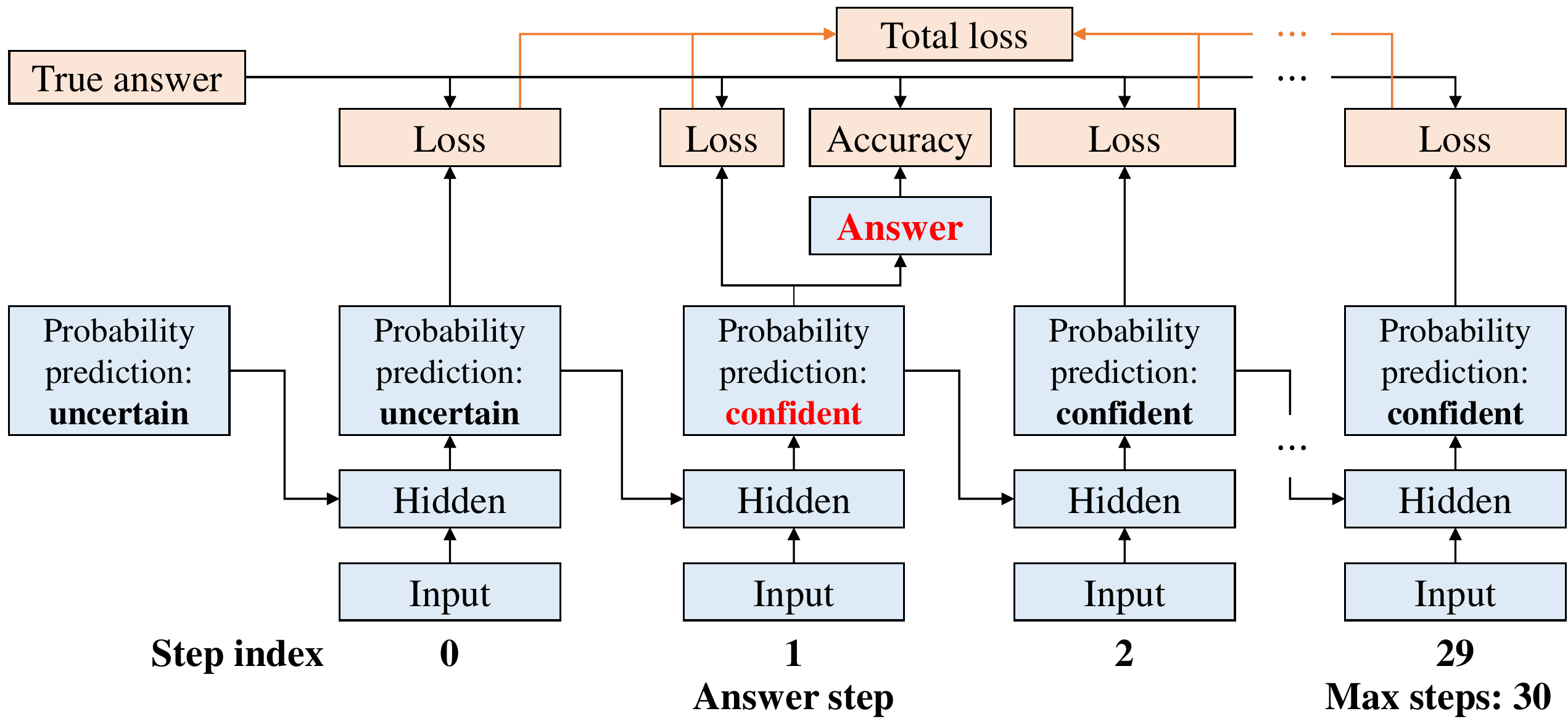}
        \caption{The Jordan network unrolled through time steps.}
        \label{fig/model_unroll_time}
    \end{subfigure}%\hfill
    \caption{The Jordan network used in the present study. (a) The network is predicting the answer of $110+1101$ to be $10011$. In this example, the confidence threshold is 0.9. At the current state $t$, $\textbf{x}^{(t)}=(0,1,1,0,1,1,0,1)$, $\textbf{p}^{(t)}=(.99, .04, .07, .96, .94)$, and $\textbf{z}^{(t)}=(1,0,0,1,1)$. (b) The network is constrained to compute at most 30 steps. The initial probabilities of answer digits are 0.5, meaning the network is uncertain about all digits. The network repeatedly computes the probabilities of answer digits until it becomes confident about all answer digits; in this figure, it answers at step 1. In the learning phase, the network learns from the total loss from all steps. Accuracy is computed by comparing predicted answers to true answers.}
    \label{fig/our_model}
\end{figure*}

\section{Experiment 2: Connectionist Models}

%Experiment 2 investigated whether connectionist models take longer time to solve a problem as it contains more carries. 
Experiment 2 investigated whether computational steps required by connectionist models in problem solving increase as a function of the number of carries involved in a problem.
Moreover, this experiment intended to examine how the central model hyperparameters --- confidence threshold and hidden dimension --- affect the simulated RT.
The hidden dimension, denoted by $d_h$, refers to the number of units in the hidden layer.

% Purpose of this experiment
% ===> Show the RNN needs cognitive process time as human participants.
% 1. Design a recurrent network that simulates human arithmetic cognition.
% 2. Empirically prove (show an existential proof) that the proposed network is able to learn addition and subtraction. (Proof of concept)
% 3. Show the RNN needs cognitive process time as human participants.

\subsection{Model}

% Motivation for modeling
Imagine the human cognitive process while performing addition and subtraction. 
Humans predict answer digits one by one while mentally referencing two operands and previously predicted digits.
% About the recurrent network model.
Therefore, we aimed to simulate this human cognitive process by using the Jordan network \cite{Jordan1997_JordanNetwork}.
The Jordan network is a recurrent neural network whose hidden layer gets its inputs from an input at the current step and from the output at the previous step (Figure \ref{fig/our_model}).

% why we come up with this neural structure.
%Before talking about the model, we should tell why we come up with this neural structure.
%% 1. We narrow the range of candidate connectionist models to the recurrent network modeling a sequence of thoughts.
%% 2. The Elman and Jordan networks were chosen as the candidates. 
%% 3. Imagine the human cognitive process while performing addition and subtraction. (At the beginning, humans are unsure of all answer digits.) Humans predict digits one by one based on previously/already predicted digits. 
%% 4. Therefore, we choose to use the Jordan network instead of the Elman network to model the human arithmetic cognition.

%The Jordan network was chosen through the following decisions.
%We first narrowed the range of connectionist models into the recurrent networks modeling a sequence of thoughts, such as the Elman and Jordan networks.

%The Jordan network performs the very process we would like to model.
%Therefore, we choose the Jordan network to model the human arithmetic cognition.

% Details about the model structure
The Jordan network solves problems as follows:
An 8-dimensional input vector composed of two concatenated 4-digit operands is fed into the network (Figure \ref{fig/jordan_network}).
At the same time, its hidden layer with ReLU gets its previous probability outputs.
The network predicts step-by-step the probabilities of answer digits up to a maximum of 30 steps (Figure \ref{fig/model_unroll_time}).
At the initial step, all digit predictions are initialized as 0.5, which mimics the initial uncertainty humans experience when solving problems. 
%At the first step, all elements of the previous probability output is initialized as 0.5 meaning that the network is uncertain about all output digits.
%Similarly, at the beginning, humans are unsure of all answer digits.
The output layer has sigmoid activation.
Each output unit predicts each output digit.
The network outputs 5-dimensional and 4-dimensional vectors for addition and subtraction problems respectively.
%The output dimension was 5 for addition and 4 for subtraction, as determined by the maximum digits of the arithmetic results.

% In the learning phase
% Backward learning policy
The network learned arithmetic by minimizing the sum of the losses at all steps: $\sum_{t}{H(\mathbf{z}^{(t)},\mathbf{p}^{(t)})}$.
At each step $t$, a loss is defined as the cross-entropy $H$ between the true answer $\mathbf{z}^{(t)}$ and the output probability vector $\mathbf{p}^{(t)}$ where $\mathbf{x}^{(t)}$ is an input vector: $H(\textbf{z}^{(t)},\textbf{p}^{(t)})=-\mathbf{z}^{(t)}\cdot\log\mathbf{p}^{(t)} - (1 - \mathbf{z}^{(t)})\cdot[1 - \log\mathbf{p}^{(t)}]$.

% In the solving phase
% About confident range and uncertain range. Define confidence thresholds.
At each time step, the network predicts the probability of every answer digit.
When problem solving, humans only decide on an answer digit when they are sufficiently confident that it is correct.
Likewise, the network decides each digit only when its predicted probability $p_i$ is higher than some threshold.
We call this threshold the \textit{confidence threshold}, denoted by $\theta_c$.
Suppose $\theta_c = 0.9$. 
If a predicted probability $p_i$ is in the range $[0.1, 0.9]$, the model is \textit{uncertain} about the digit. 
Otherwise, it is \textit{confident} about the digit: if $p_i \in [0,0.1)$, it predicts the digit is 0; if $p_i \in (0.9,1]$, it predicts the digit is 1.
The network is designed to give an \textit{answer} when it is first confident about all answer digits (Figure \ref{fig/model_unroll_time}). 
The network in Figure \ref{fig/model_unroll_time} answers at step 1 because this is the first state where the model is confident about all digits.
%At this answer step, the correctness of the answer is determined.
At this answer step, the answer is marked as either correct or incorrect.
No answer is given if 30 steps are exceeded.

\subsection{Measures}

%The two following measures are notable in this experiment.

\subsubsection{Accuracy}
% #(correct_answers) / #(problems)
% #(correct_answers) = #(wrong answers) + #(none answers)
\textit{Accuracy} was measured by dividing the number of correct answers by the total number of problems.
Model accuracy was used to measure how successfully the model learned arithmetic and to determine when to stop training.
No answer after 30 time steps was considered a wrong answer.

\subsubsection{Answer step}
\textit{Answer step} was defined as the index of a certain time step where the network outputs an answer. 
Answer step is roughly equivalent to human RT.
It refers to the number of computational steps required for the network to solve an arithmetic problem. 
Answer step ranges from 0 to 29.
%To measure computational steps of a particular problem set, we observed the mean answer step of the set.

\subsection{Training Settings}

The network learned arithmetic operations using backpropagation through time \cite{Werbos1990_BPTT} and a stocbohastic gradient method \cite{Bottou98} called Adam optimization \cite{KingmaB14} with settings ($\alpha=.001$, $\beta_{1}=.9$, $\beta_{2}=.999$, $\epsilon=10^{-8}$). 
For each epoch, 32-sized mini-batches were randomly sampled without replacement \cite{Shamir16} from the total operation dataset. 
The weight matrix $W^{[l]}$ in layer $l$ was initialized to samples from the truncated normal distribution ranging $[-{1}/{\sqrt{n^{[l-1]}}},{1}/{\sqrt{n^{[l-1]}}}]$ where $n^{[l]}$ was the number of units in the $l$-th layer; All bias vectors $b^{[l]}$ were initialized to 0.
After training each epoch, accuracy was evaluated on the operation dataset (Figure \ref{fig/problem_sets}). 
When the network attained 100\% accuracy for the entirety of the operation dataset, training was stopped. 
%In this experiment, we were primarily concerned about the increasing relationship between mean answer step and the number of carries involved in arithmetic problems. 
300 Jordan networks were trained for each model configuration in order to draw statistically meaningful results.
Furthermore, to investigate if any statistically significant relationship held for various model configurations, we reanalyzed the models with the confidence thresholds $\theta_{c} \in \{.7,.8,.9\}$ and hidden dimensions $d_h \in \{24,48,72\}$.
9 types of networks were trained for both addition and subtraction, respectively; a total of 5400 networks were trained in this experiment.

\subsection{Results}

Our proposed model successfully learned all possible addition and subtraction operations between 4-digit binary numbers.
The model required 4000 epochs on average (58 minutes\footnote{Two Intel(R) Xeon(R) CPU E5-2695 v4 and five TITAN Xp were used. Training networks in parallel is vital in this experiment.}) to learn addition, and 1080 epochs on average (13 minutes) to learn subtraction.
%
%it had learned to correctly solve all the possible problems, i.e., 
When training was completed, we examined:
(1) statistical differences in mean answer steps between carry datasets across all model configurations;
(2) statistical differences in mean answer steps for operation datasets between different confidence thresholds and hidden dimensions.
%ANOVA was used to check for significant differences between all mean answer steps. 
%If differences were found, post hoc analysis was applied.

\subsubsection{Addition}

The first analysis was conducted on mean answer steps per carry dataset.
For every model configuration, ANOVA found significant differences in mean answer steps between all carry datasets (Table \ref{table/mean_answer_step_by_carries}). 
Post hoc Games-Howell testing found that for 8 of the 9 model configurations, mean answer step was strictly increasing with respect to the number of carries (Table \ref{table/mean_answer_step_by_carries}, Figure \ref{fig/mean_answer_step_by_carries_add}); the remaining model configuration ($\theta_c = 0.7$, $d_h = 24$) showed a monotonically\footnote{For every $x$ and $x'$ such that $x<x'$, if $f(x) \le f(x')$, then we say $f$ is \textit{monotonically increasing}.} increasing relationship between mean answer step and the number of carries (Table \ref{table/mean_answer_step_by_carries}).

%Further, we examine how confidence threshold and the hidden dimension of the models affect mean answer step for the addition dataset.

The second analyses were conducted on mean answer steps for the addition dataset.
For every hidden dimension, ANOVA found significant differences in mean answer steps between all confidence thresholds $^\forall \theta_{c} \in \{.7,.8,.9\}$ (Table \ref{table/mean_answer_step_by_confidence_probability}).
Post hoc Games-Howell testing found that for all models, mean answer step was strictly increasing with respect to confidence threshold (Table \ref{table/mean_answer_step_by_confidence_probability}, Figure \ref{fig/mean_answer_step_by_confidence_probability_add}).
For every confidence threshold, ANOVA found significant differences in mean answer steps between all hidden dimensions $^\forall d_h \in \{24,48,72\}$ (Table \ref{table/mean_answer_step_by_hidden_dimension}).
Post hoc Games-Howell testing found that with $\theta_c = 0.7$, mean answer step was monotonically increasing with respect to  hidden dimension. For both other confidence thresholds, mean answer step was strictly increasing with respect to hidden dimension (Table \ref{table/mean_answer_step_by_hidden_dimension}, Figure \ref{fig/mean_answer_step_by_hidden_dimension_add}).
We should note however that while significant, the effect of hidden dimension on mean answer step was small.

\subsubsection{Subtraction}

\begin{figure}[bt!]
    \centering
    \begin{subfigure}[b]{0.25\textwidth}
        \centering
        \includegraphics[height=4.3cm]{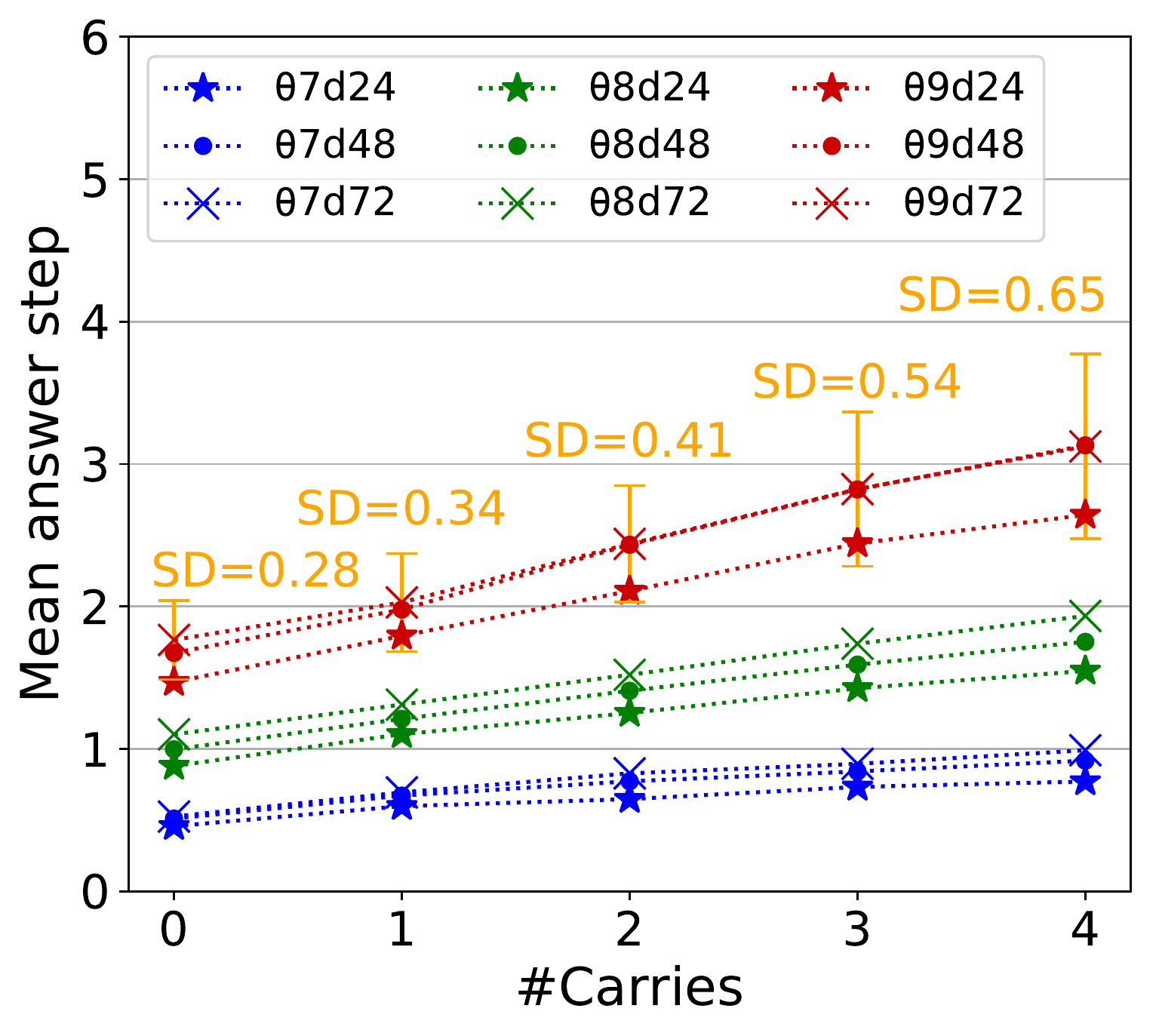}
        \caption{Addition}
        \label{fig/mean_answer_step_by_carries_add}
    \end{subfigure}%\hfill
    \begin{subfigure}[b]{0.25\textwidth}
        \centering
        \includegraphics[height=4.3cm]{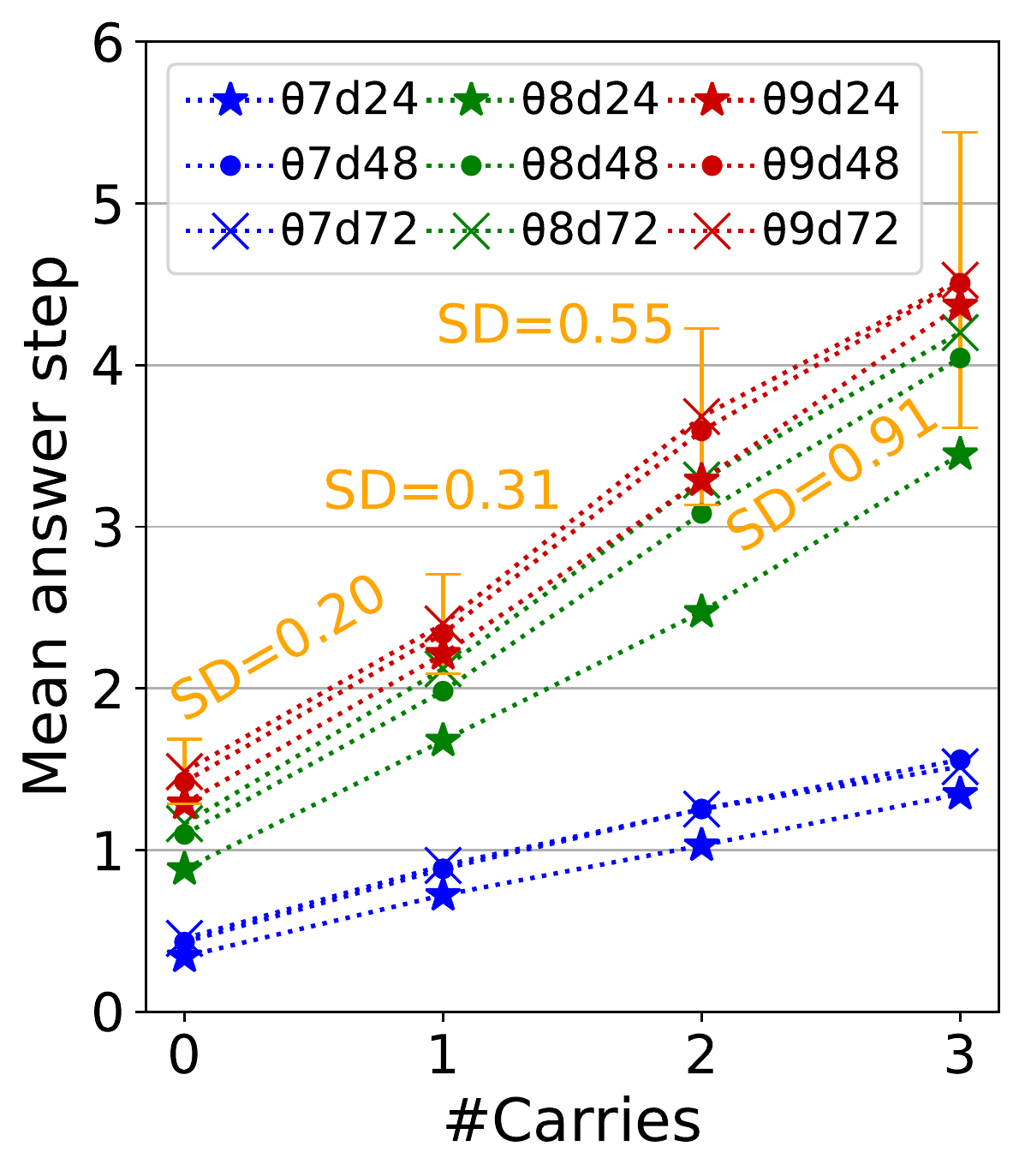}
        \caption{Subtraction}
        \label{fig/mean_answer_step_by_carries_subtract}
    \end{subfigure}%\hfill
    \caption{Mean answer step by carries (for carry datasets). $\theta$9d72 denotes models with $\theta_c = 0.9$ and $d_h=72$. The error bars are $\pm 1SD$ and belong to $\theta$9d72.}
    \label{fig/mean_answer_step_by_carries}
\end{figure}

The first analysis was conducted on mean answer steps per carry dataset.
For every model configuration, ANOVA found significant differences in mean answer steps between all carry datasets (Table \ref{table/mean_answer_step_by_carries}). 
Post hoc Games-Howell testing found that for all model types, mean answer step was strictly increasing with respect to the number of carries (Table \ref{table/mean_answer_step_by_carries}, Figure \ref{fig/mean_answer_step_by_carries_subtract}).

%For every model type, there were significant differences between all mean answer steps of carry datasets, as determined by ANOVA (Table \ref{table/mean_answer_step_by_carries}).
%Post hoc comparisons using the Games-Howell test indicated that all 9 types of the Jordan networks showed a strictly increasing tendency between mean answer step and carries (Table \ref{table/mean_answer_step_by_carries}, Figure \ref{fig/mean_answer_step_by_carries_subtract}).

\begin{figure}[bt!]
    \centering
    \begin{subfigure}[b]{0.25\textwidth}
        \centering
        \includegraphics[width=\textwidth]{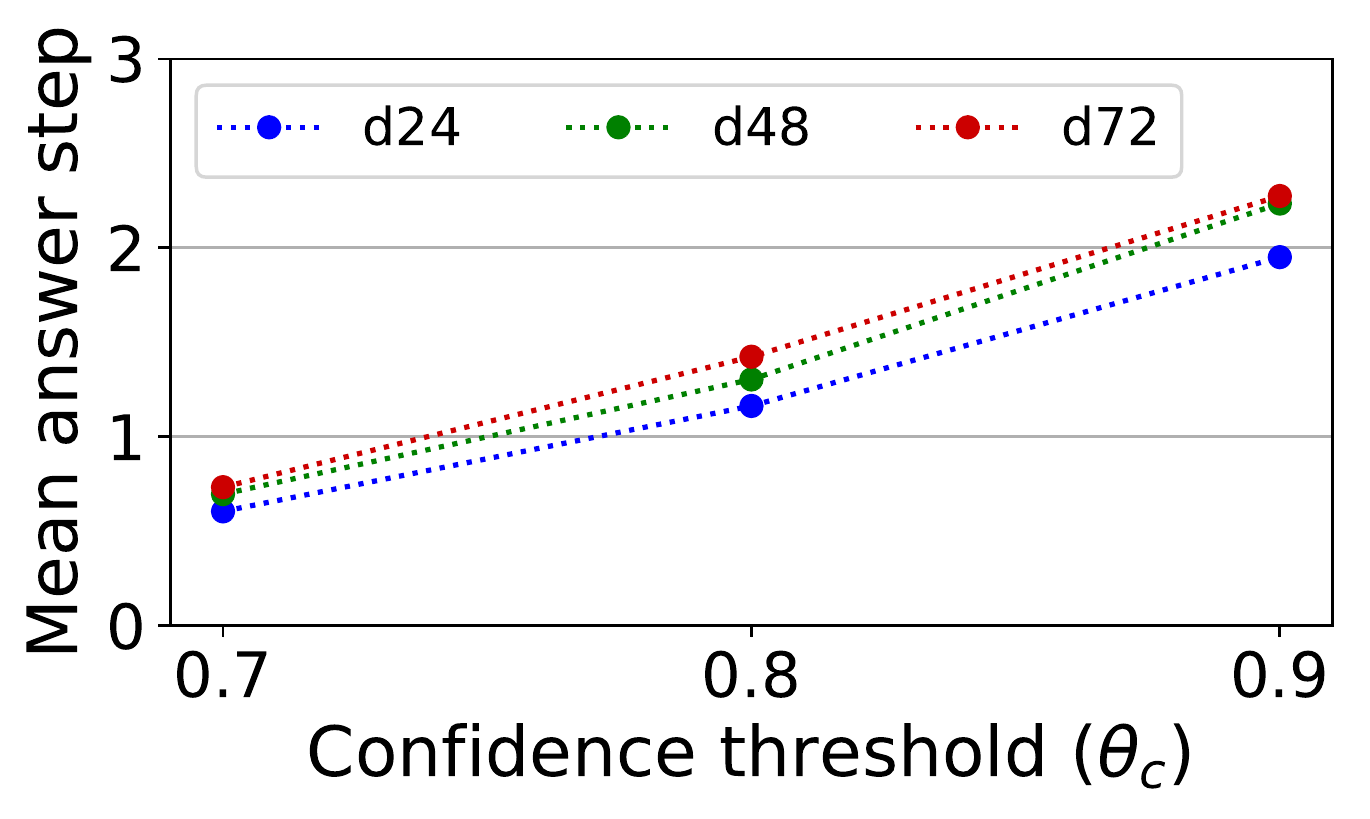}
        \caption{Addition}
        \label{fig/mean_answer_step_by_confidence_probability_add}
    \end{subfigure}%\hfill
    \begin{subfigure}[b]{0.25\textwidth}
        \centering
        \includegraphics[width=\textwidth]{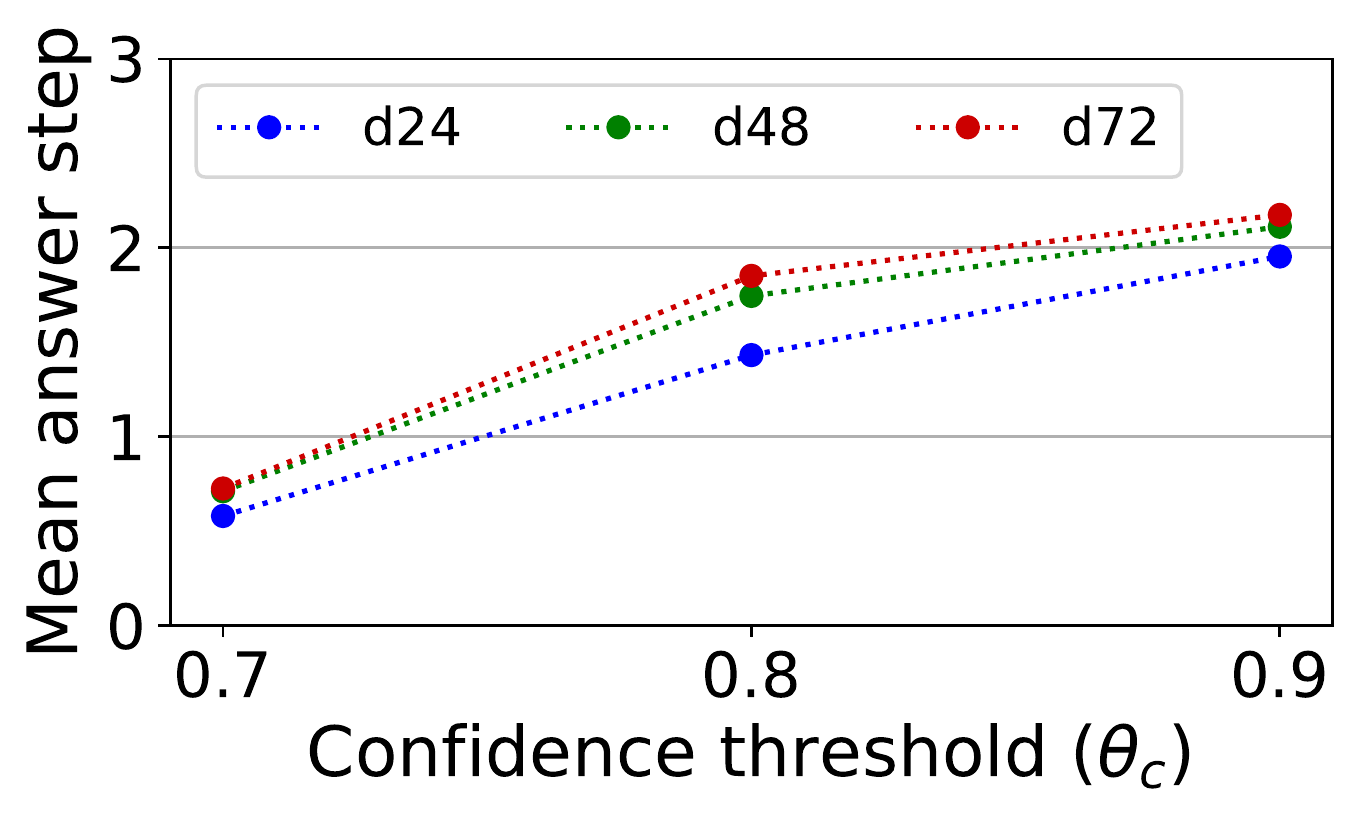}
        \caption{Subtraction}
        \label{fig/mean_answer_step_by_confidence_probability_subtract}
    \end{subfigure}%\hfill
    \caption{Mean answer step by confidence threshold (for operation datasets)}
    \label{fig/mean_answer_step_by_confidence_probability}
\end{figure}

\begin{figure}[bt!]
    \centering
    \begin{subfigure}[b]{0.25\textwidth}
        \centering
        \includegraphics[width=\textwidth]{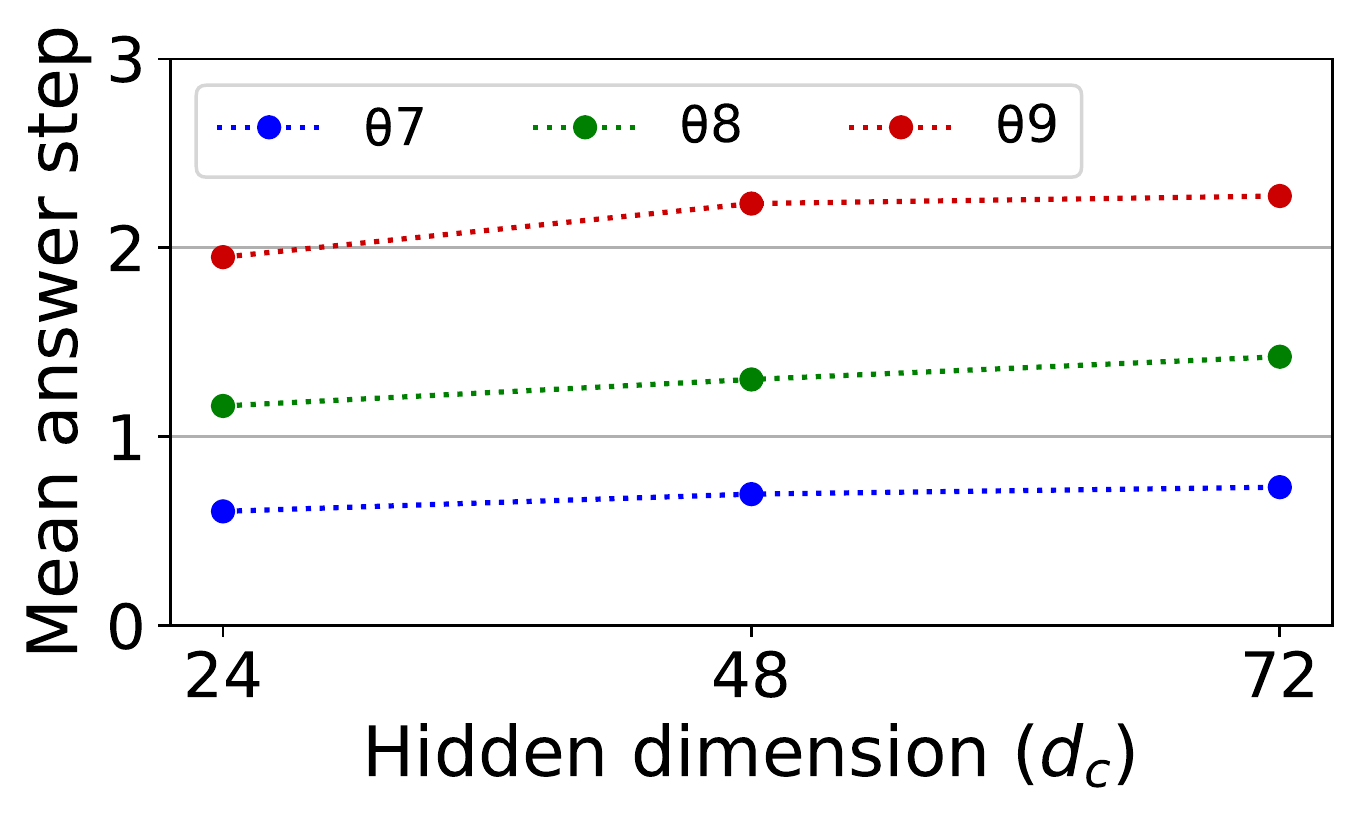}
        \caption{Addition}
        \label{fig/mean_answer_step_by_hidden_dimension_add}
    \end{subfigure}%\hfill
    \begin{subfigure}[b]{0.25\textwidth}
        \centering
        \includegraphics[width=\textwidth]{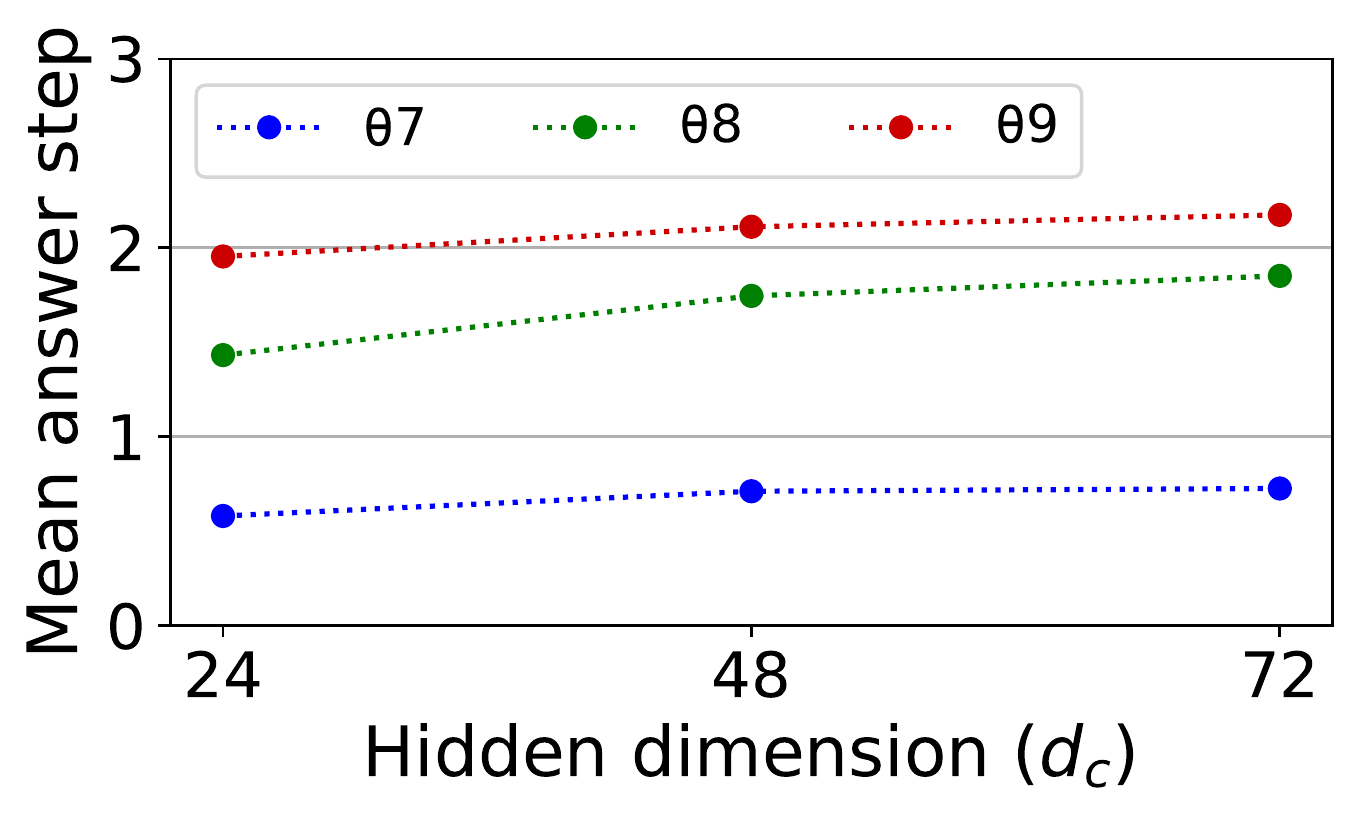}
        \caption{Subtraction}
        \label{fig/mean_answer_step_by_hidden_dimension_subtract}
    \end{subfigure}%\hfill
    \caption{Mean answer step by hidden dimension (for operation datasets)}
    \label{fig/mean_answer_step_by_hidden_dimension}
\end{figure}

The second analyses were conducted on mean answer steps for the subtraction dataset.
For every hidden dimension, ANOVA found significant differences in mean answer steps between all confidence thresholds $^\forall \theta_{c} \in \{.7,.8,.9\}$ (Table \ref{table/mean_answer_step_by_confidence_probability}).
Post hoc Games-Howell testing found that for all models, mean answer step was strictly increasing with respect to confidence threshold (Table \ref{table/mean_answer_step_by_confidence_probability}, Figure \ref{fig/mean_answer_step_by_confidence_probability_subtract}).
%For every hidden dimension, there were significant differences between all mean answer steps of confidence probabilities --- 0.7, 0.8, and 0.9 --- as determined by ANOVA (Table \ref{table/mean_answer_step_by_confidence_probability}).
%Post hoc comparisons using the Games-Howell test indicated that the networks showed a strictly increasing tendency between mean answer step and confidence thresholds (Table \ref{table/mean_answer_step_by_confidence_probability}, Figure \ref{fig/mean_answer_step_by_confidence_probability_subtract}); this implies high confidence thresholds made the model take more steps to arrive at the correct answer.
For every confidence threshold, ANOVA found significant differences in mean answer steps between all hidden dimensions $^\forall d_h \in \{24,48,72\}$ (Table \ref{table/mean_answer_step_by_hidden_dimension}).
Post hoc Games-Howell testing found that with $\theta_c = 0.9$, mean answer step was monotonically increasing with respect to hidden dimension. For both other confidence thresholds, mean answer step was strictly increasing with respect to hidden dimension (Table \ref{table/mean_answer_step_by_hidden_dimension}, Figure \ref{fig/mean_answer_step_by_hidden_dimension_add}).
We should note however that while significant, the effect of hidden dimension on mean answer step was small (Figure \ref{fig/mean_answer_step_by_hidden_dimension_add}).
%For every confidence threshold, there were significant differences between all mean answer steps of hidden dimensions --- 24, 48, and 72 --- as determined by ANOVA (Table \ref{table/mean_answer_step_by_hidden_dimension}).
%Post hoc comparisons using the Games-Howell test indicated that the networks showed a monotonically increasing tendency between mean answer step and hidden dimension for the confidence threshold 0.9 but a strictly increasing tendency for the others (Table \ref{table/mean_answer_step_by_hidden_dimension}, Figure \ref{fig/mean_answer_step_by_hidden_dimension_subtract}); this implies larger hidden dimension tend to make the model take more steps to arrive at the correct answer. 
%However, hidden dimension very slightly affect mean answer step (Figure \ref{fig/mean_answer_step_by_hidden_dimension_subtract}).

\begin{table}[bt!]
\setlength\tabcolsep{3.8pt}
%\footnotesize
\scriptsize
\begin{center} 
\caption{The results of ANOVA and post hoc analysis on differences in mean answer steps between all carry datasets. The model configuration varies along two axes: confidence threshold and hidden dimension. 300 mean answer steps per carry dataset from 300 trained networks were analyzed for each model configuration. $F$ is the $F$-test statistic and $\eta^2$ is the effect size from ANOVA; in addition, there were 4 degrees of freedom between carry datasets and 1495 within carry datasets: $\textup{df}_{b}^{+}=4$,  $\textup{df}_{w}^{+}=1495$; in subtraction, $\textup{df}_{b}^{-}=3$,  $\textup{df}_{w}^{-}=1196$. The mean answer step columns describe the results of post hoc analysis. The inequality ($<$) denotes a significant difference at the $p<.05$ level. Equality ($=$) denotes the opposite. The numbers in these columns refer to the number of carries of a carry dataset. 
%The following symbols indicate confidence level of the results. 
$^{*}$ $p<.05$. $^{**}$ $p<.01$. $^{***}$ $p<.001$.} 
\label{table/mean_answer_step_by_carries} 
\vskip 0.12in
\begin{tabular}{cc|rl|l|rl|l}
\toprule
 &  & \multicolumn{3}{c}{\textbf{Addition}} & \multicolumn{3}{|c}{\textbf{Subtraction}}   \\
%\midrule
%& & \multicolumn{2}{c}{ANOVA} & \multicolumn{1}{|c|}{Post hoc} & \multicolumn{2}{c}{ANOVA}  & \multicolumn{1}{|c}{Post hoc} \\
\midrule
\textbf{$\theta_c$} & \textbf{$d_h$} & \multicolumn{1}{c}{$F$} & \multicolumn{1}{c}{$\eta^2$} & \multicolumn{1}{|c|}{Mean answer step} & \multicolumn{1}{c}{$F$} & \multicolumn{1}{c}{$\eta^2$} & \multicolumn{1}{|c}{Mean answer step} \\
\midrule
.7     & 24 & $72^{***}$   &.16& $0 < 1 = 2 < 3 = 4^{***}$ & $499^{***}$   &.56& $0 < 1 < 2 < 3^{***}$     \\
.7     & 48 & $206^{***}$  &.36& $0 < 1 < 2 < 3 < 4^{***}$ & $765^{***}$   &.66& $0 < 1 < 2 < 3^{***}$    \\
.7     & 72 & $294^{***}$  &.44& $0 < 1 < 2 < 3 < 4^{***}$ & $716^{***}$   &.64& $0 < 1 < 2 < 3^{***}$    \\
\midrule
.8     & 24 & $129^{***}$  &.26& $0 < 1 < 2 < 3 < 4^{**}$  & $390^{***}$   &.49& $0 < 1 < 2 < 3^{***}$     \\
.8     & 48 & $198^{***}$  &.35& $0 < 1 < 2 < 3 < 4^{***}$ & $571^{***}$   &.59& $0 < 1 < 2 < 3^{***}$    \\
.8     & 72 & $142^{***}$  &.28& $0 < 1 < 2 < 3 < 4^{**}$  & $674^{***}$   &.63& $0 < 1 < 2 < 3^{***}$    \\
\midrule
.9     & 24 & $208^{***}$  &.36& $0 < 1 < 2 < 3 < 4^{**}$  & $970^{***}$   &.71& $0 < 1 < 2 < 3^{***}$     \\
.9     & 48 & $421^{***}$  &.53& $0 < 1 < 2 < 3 < 4^{***}$ & $1769^{***}$  &.82& $0 < 1 < 2 < 3^{***}$    \\
.9     & 72 & $432^{***}$  &.54& $0 < 1 < 2 < 3 < 4^{***}$ & $1718^{***}$  &.81& $0 < 1 < 2 < 3^{***}$    \\
\bottomrule
\end{tabular}
\end{center}
\end{table}

\begin{table}[t!]
\setlength\tabcolsep{3.8pt}
%\footnotesize
\scriptsize
\begin{center} 
\caption{The results of ANOVA and post hoc analysis on differences in mean answer steps between confidence thresholds. 
%300 mean answer steps from 300 trained models were analyzed for each model configuration.
$\textup{df}_{b}^{+}=\textup{df}_{b}^{-}=2$.  $\textup{df}_{w}^{+}=\textup{df}_{w}^{-}=897$.
In the mean answer step columns, the numbers refer to confidence thresholds.
%The symbols used in this table are consistent with those in Table \ref{table/mean_answer_step_by_carries}. 
} 
\label{table/mean_answer_step_by_confidence_probability} 
\vskip 0.12in
\begin{tabular}{c|rl|l|rl|l}
%\begin{tabularx}{\textwidth}{c|rl|l|rl|l}
\toprule
& \multicolumn{3}{c}{\textbf{Addition}} & \multicolumn{3}{|c}{\textbf{Subtraction}}   \\
%\midrule
%& \multicolumn{2}{c}{ANOVA} & \multicolumn{1}{|c|}{Post hoc} & \multicolumn{2}{c}{ANOVA}  & \multicolumn{1}{|c}{Post hoc} \\
\midrule
\textbf{$d_h$} & \multicolumn{1}{c}{$F$} & \multicolumn{1}{c}{$\eta^2$} & \multicolumn{1}{|c|}{Mean answer step} & \multicolumn{1}{c}{$F$} & \multicolumn{1}{c}{$\eta^2$} & \multicolumn{1}{|c}{Mean answer step} \\
\midrule
24  & $1032^{***}$ &.70& $.7 < .8 < .9^{***}$    & $1163^{***}$  &.72& $.7 < .8 < .9^{***}$    \\
48  & $2002^{***}$ &.82& $.7 < .8 < .9^{***}$    & $1736^{***}$  &.79& $.7 < .8 < .9^{***}$    \\
72  & $1735^{***}$ &.79& $.7 < .8 < .9^{***}$    & $1963^{***}$  &.81& $.7 < .8 < .9^{***}$    \\
\bottomrule
\end{tabular}
%\end{tabularx}
\end{center}
\end{table}

\begin{table}[t!]
%\footnotesize
\scriptsize
\begin{center}
\setlength\tabcolsep{3.8pt}
\caption{The results of ANOVA and post hoc analysis on differences in mean answer steps between hidden dimensions. 
%300 mean answer steps from 300 trained models were analyzed for each model configuration.
$\textup{df}_{b}^{+}=\textup{df}_{b}^{-}=2$.  $\textup{df}_{w}^{+}=\textup{df}_{w}^{-}=897$.
In the mean answer step columns, the numbers refer to hidden dimension.
%The symbols used in this table are consistent with those in Table \ref{table/mean_answer_step_by_carries}.  
} 
\label{table/mean_answer_step_by_hidden_dimension} 
\vskip 0.12in
\begin{tabular}{c|rl|l|rl|l}
\toprule
 & \multicolumn{3}{c}{\textbf{Addition}} & \multicolumn{3}{|c}{\textbf{Subtraction}}   \\
%\midrule
%& \multicolumn{2}{c}{ANOVA} & \multicolumn{1}{|c|}{Post hoc} & \multicolumn{2}{c}{ANOVA}  & \multicolumn{1}{|c}{Post hoc} \\
\midrule
\textbf{$\theta_c$} & \multicolumn{1}{c}{$F$} & \multicolumn{1}{c}{$\eta^2$} & \multicolumn{1}{|c|}{Mean answer step} & \multicolumn{1}{c}{$F$} & \multicolumn{1}{c}{$\eta^2$} & \multicolumn{1}{|c}{Mean answer step} \\
\midrule
.7  & $58^{***}$ &.08& $24 < 48 = 72^{***}$   & $46^{***}$ &.10& $24 < 48 < 72^{**}$    \\
.8  & $38^{***}$ &.08& $24 < 48 < 72^{***}$   & $77^{***}$ &.15& $24 < 48 < 72^{**}$    \\
.9  & $37^{***}$ &.12& $24 < 48 < 72^{*}$     & $51^{***}$ &.09& $24 < 48 = 72^{***}$   \\
\bottomrule
\end{tabular}
\end{center}
\end{table}

\section{Discussion and Conclusion}
% Why the discussion part is needed.
% https://library.sacredheart.edu/c.php?g=29803&p=185933
% The purpose of the discussion is to interpret and describe the significance of your findings in light of what was already known about the research problem being investigated, and to explain any new understanding or fresh insights about the problem after you've taken the findings into consideration.
% The discussion will always connect to the introduction by way of the research questions or hypotheses you posed and the literature you reviewed, but it does not simply repeat or rearrange the introduction; 
% the discussion should always explain how your study has moved the reader's understanding of the research problem forward from where you left them at the end of the introduction.
%
% the underlying meaning of your research
% present the importance of your study
% Use the present verb tense, especially for established facts; however, refer to specific works and references in the past tense.

% Summary the results very abstract level.

\subsubsection{Experiment 1}
Experiment 1 has improved the previous study \cite{Cho2019} as follows: 
Firstly, participants were forced to solve problems using solely mental arithmetic. 
This allows for more valid comparisons to be drawn between humans and models.
%this study forced participants to solve problems using solely mental arithmetic for comparing fairly those with connectionist  models  without  external  memory
%(b) it was found that mean RT for addition problems was strictly increasing with respect to the number of carries.
Secondly, larger data samples allowed the present study to find more statistically significant results. Specifically, mean RT for addition problems were found to be strictly increasing with respect to the number of carries.

\subsubsection{Experiment 2}
In Experiment 2, the two hyperparameters --- confidence threshold and hidden dimension --- were chosen since we expected these hyperparameters to correspond to humans' uncertainty and memory capacity, respectively. 
We further expected that increasing confidence threshold and decreasing hidden dimension would increase answer step. 
This expectation subsequently arose for confidence threshold; confidence threshold had an augmenting effect on answer step.
However, our expectation was not born out for hidden dimension.
%Consequently, these two variables make the model take more steps to answer; the former certainly contributes to answer step but the latter dose not. 
In order to observe clear differences in mean answer steps with respect to problem difficulty, high confidence thresholds are recommended.
%It is enough to fix hidden dimension as much as the model can learn a total dataset.
Hidden dimension should be fixed to the extent that the model can learn an entire dataset. 
%Large hidden dimension induces the model to learn a problem set in short time. 

%Hidden dimension were expected to be major factor to make answer steps vary depending on problem difficulty. Because we thought less capacity made varying answer steps. But it is not. 

%max step increases learning complexity. less than 30 seems work.  
\subsubsection{Experiments 1 \& 2}
The preceding results show three notable similarities between humans and our connectionist models:
Firstly, both agents experienced increased levels of difficulty as more carries were involved in arithmetic problems.
Secondly, the Jordan networks with the model configuration ($\theta_{c} = 0.9$, $d_h = 72$) successfully mimicked the increasing standard deviation of human RT with respect to the number of carries (Figure \ref{fig/response_time_by_carries}, \ref{fig/mean_answer_step_by_carries}).
This phenomenon could not be achieved by a rule-based system performing the standard algorithm, although such a system would be able to simulate increasing RT as a function of the number of carries.
Lastly, another similarity found between both humans and models is that the difficulty slope for subtraction is steeper than for addition (Figure \ref{fig/response_time_by_carries}, \ref{fig/mean_answer_step_by_carries}). This implies that the augmenting effect of carries on problem difficulty is stronger in subtraction than in addition. 

\subsubsection{Contributions}
The present study makes two major contributions to the literature:
%Firstly, the present study demonstrates a connectionist model that experiences difficulty on arithmetic problems like humans.
%Firstly, the present study directly compares both humans and connectionist models on arithmetic tasks.
Firstly, our models successfully simulated humans' RT in terms of these three similarities: increasing latency, increasing standard deviation of latency, and relative steepness of increasing latency.
The similarities may suggest that some cognitive process, equivalent to the nonlinear computational process used in the Jordan network, could be involved in human cognitive arithmetic.
Secondly, the present study demonstrated that fitting our model to arithmetic data induced human-like latency to emerge in the connectionist models \cite{Mcclelland2010_letting}.
In other words, human RTs to arithmetic problems were successfully learned in an unsupervised way.
This contrasts with previous studies that focus on learning arithmetic tasks in a supervised way.
%Fourthly, our connectionist models provide a concrete instantiation of the distributed processing mechanisms \cite{RumelhartM86} that grasp arithmetic cognition.
%Finally, the fact that the model requires more computation to solve difficult problems supports the principle that many nonlinear computational steps are required to learn complex mappings \cite{Lecun2015_deepLearning}.

\subsubsection{Future Study}

%Arithmetic was chosen as the domain of cognition to simulate by our connectionist model. This is because the symbolic representation and processing involved in cognitive arithmetic makes this process more explicitly observable than other cognitive domains.
%Arithmetic was chosen as the domain of cognition to simulate our connectionist model because the cognitive process of arithmetic is be more explicitly observable than other domains, due to its symbolic representation and processing.
%but the dynamic processing of individual problems has not been explored yet.
The present study focuses solely on analyzing mean answer steps between arithmetic problem sets of varying difficulty levels. 
Therefore, future studies could aim to better understand what dynamic processes our model uses when solving individual problems: Specifically, it might be interesting to observe how our model predicts individual digits through each time step when solving problems.
Furthermore, similarities between both the model's sequentially predictive answering process and the human answering process could be investigated.
%Furthermore, how its sequential predictions are similar to humans' solution should be explored. 
%This will give us better understanding on the model mechanism.  
This comparison would give us a better understanding of both our model and human mathematical cognition \cite{McclellandMHYL16}. %  for examples, why mathematics is hard to learn

\vspace{6 pt}
%Simulating response time using connectionist models: rare case. 
Our model is designed not just for arithmetic cognition, but also for sequential predictions that based on a constant input and a previous prediction, which result in a single answer.
In this regard, this model has the potential to be applied to other cognitive processes involving sequential processing and RT as a measure of cognitive difficulty.
Therefore, future studies could consider extending our model to other domains of cognition.
For example, well known character image and word classification datasets can be subdivided into datasets of varying difficulty levels, similar to our carry datasets. 
Mean answer steps for classifying these data sets could be analyzed using a similar model to that outlined in the present study.
%For example, the followings could be further investigated: mean answer times between character images easy to recognize and those hard to recognize, or mean answer times between words meaning easy concept and those meaning hard concept.

% [similar phenomenon. Side ]
%When problems have the neighboring number of carries, it is a high chance to meet the case.
%With small samples, the mean answer steps of carry problems that have difference more than one carries are significantly different. 
%However, carry problems that have one carry of difference need more samples to show significant difference.

%[The Significance of PDP Approach in Mathematical Cognition] 
%According to \citeA{McclellandMHYL16}, extending the connectionist approach \cite{RumelhartM86} to address problems of mathematical cognition may help us understand in more detail why mathematics is hard to learn. 
%This approach is effective because connectionist models are able to learn many aspects of mathematical cognition. 

%\citeA{Mickey2014} showed that their connectionist model is able to simulate several phenomena observed in how children learn mathematical equivalence.
%Recently, mathematics datasets to test extensive mathematical skills of neural models have been released \cite{Saxton2018_mathDataset}.

% At the early stage, accuracy = 0. Many none answers 
% Then, it tends to reduce mean answer step of overall problem solving.

\section{Acknowledgments}
%Joonho Kim, Chung-Yeon Lee, Kyoung-won on
We thank
Seho Park, Seung Hee Yang, Chung-Yeon Lee, Gi-Cheon Kang, and Paula Higgins for useful discussion and writing comments.
%This work was partly supported by the Institute for Information \& Communications Technology Promotion (R0126-16-1072-SW.StarLab, 2017-0-01772-VTT, 2018-0-00622-RMI, 2019-0-01367-BabyMind) and Korea Evaluation Institute of Industrial Technology (10060086-RISF) grant funded by the Korea government (MSIP, DAPA).
%This work was partly supported by the Korea government (IITP-R0126-16-1072-SW.StarLab, IITP-2017-0-01772-VTT, IITP-2018-0-00622-RMI, IITP-2019-0-01367-BabyMind, KEIT-10060086-RISF).
This work was supported by a grant to Biomimetic Robot Research Center funded by Defense Acquisition Program Administration and Agency for Defense Development (UD160027ID).

% Adding references without citing

%\nocite{HoshenP16}
%\nocite{Feigenbaum1963a}
%\nocite{Hill1983a}
%\nocite{OhlssonLangley1985a}
% \nocite{Lewis1978a}
%\f{Matlock2001}

\bibliographystyle{apacite}

\setlength{\bibleftmargin}{.125in}
\setlength{\bibindent}{-\bibleftmargin}

\bibliography{CogSci_Template}

\newpage

\onecolumn
\section{Appendix}

\begin{table*}[!ht]
\footnotesize
\centering
\caption{Means (and standard deviations) of mean RTs in Experiment 1}
\begin{tabular}{cccccc} 
\toprule
\multirow{2}{*}{\bfseries Operator} & \multicolumn{5}{c}{\textbf{Carries}} \\
& \bfseries 0 & \bfseries 1 & \bfseries 2 & \bfseries  3 & \bfseries 4 \\  
\midrule
\bfseries Addition   & 3.81  & 4.29  & 4.75  & 5.43  & 6.11   \\
                     & (0.69)& (0.88)& (0.94)& (1.25)& (1.86) \\
\addlinespace[0.2cm]
\bfseries Subtraction & 3.46  & 5.04  & 6.85  & 8.46  &   \\
                      & (0.68)& (1.45)& (2.05)& (2.78)&    \\
\bottomrule
\addlinespace[0.2cm]
\multicolumn{6}{r}{($n=90$ for each group)}
\end{tabular}
\end{table*}

\begin{table*}[!ht]
\footnotesize
\centering
\caption{Means (and standard deviations) of mean answer steps in Experiment 2}
\begin{tabular}{ccccccccc} 
\toprule
\multirow{2}{*}{\bfseries Operator} & \multirow{2}{*}{$\theta_c$} &  \multirow{2}{*}{$n_h$} & \multirow{2}{*}{\bfseries All} & \multicolumn{5}{c}{\bfseries Carries} \\
&  &  & &  \bfseries 0 & \bfseries 1 & \bfseries 2 & \bfseries  3 & \bfseries 4 \\ 
\midrule
& \multirow{2}{*}{.7} & \multirow{2}{*}{24} & 0.60  & 0.46 & 0.60 & 0.65 & 0.73 & 0.77 \\
&  &                                         & (0.23) & (0.25) & (0.28) & (0.25) & (0.24) & (0.24)  \\
\addlinespace[0.2cm]
& \multirow{2}{*}{.7} & \multirow{2}{*}{48} & 0.70  & 0.51 & 0.67 & 0.77 & 0.84 & 0.92 \\
& &                                         & (0.16) & (0.19) & (0.19) & (0.18) & (0.17) & (0.21) \\
\addlinespace[0.2cm]
& \multirow{2}{*}{.7} & \multirow{2}{*}{72} & 0.73 & 0.52 & 0.69 & 0.83 & 0.89 & 0.99 \\
& &                                         & (0.16) & (0.20) & (0.19) & (0.16) & (0.16) & (0.20) \\
\addlinespace[0.2cm]
& \multirow{2}{*}{.8} & \multirow{2}{*}{24} & 1.16 & 0.88 & 1.10 & 1.25 & 1.43 & 1.55 \\
&  &                                        & (0.36) & (0.38) & (0.40) & (0.37) & (0.38) & (0.47)  \\
\addlinespace[0.2cm]
\multirow{2}{*}{\bfseries Addition}
& \multirow{2}{*}{.8} & \multirow{2}{*}{48} & 1.30 & 1.00 & 1.21 & 1.41 & 1.59 & 1.75 \\
&  &                                        & (0.32) & (0.32) & (0.31) & (0.34) & (0.38) & (0.47)  \\
\addlinespace[0.2cm]
& \multirow{2}{*}{.8} & \multirow{2}{*}{72} & 1.42 & 1.10 & 1.31 & 1.52 & 1.74 & 1.93 \\
&  &                                        & (0.41) & (0.33) & (0.34) & (0.45) & (0.55) & (0.64)  \\
\addlinespace[0.2cm]
& \multirow{2}{*}{.9} & \multirow{2}{*}{24} & 1.95 & 1.47 & 1.79 & 2.11 & 2.44 & 2.64 \\
&  &                                        & (0.47) & (0.45) & (0.47) & (0.51) & (0.63) & (0.74)  \\
\addlinespace[0.2cm]
& \multirow{2}{*}{.9} & \multirow{2}{*}{48} & 2.23 & 1.67 & 1.98 & 2.43 & 2.82 & 3.13 \\
&  &                                        & (0.38) & (0.34) & (0.38) & (0.46) & (0.60) & (0.66)  \\
\addlinespace[0.2cm]
& \multirow{2}{*}{.9} & \multirow{2}{*}{72} & 2.27 & 1.76 & 2.03 & 2.44 & 2.82 & 3.12 \\
&  &                                        & (0.34) & (0.28) & (0.34) & (0.41) & (0.54) & (0.65)  \\
\addlinespace[0.2cm]
\midrule
& \multirow{2}{*}{.7} & \multirow{2}{*}{24} & 0.58 & 0.34 & 0.72 & 1.03 & 1.35 &  \\
&  &                                        & (0.21) & (0.17) & (0.27) & (0.35) & (0.47) &  \\
\addlinespace[0.2cm]
& \multirow{2}{*}{.7} & \multirow{2}{*}{48} & 0.71 & 0.43 & 0.88 & 1.25 & 1.56 & \\
&  &                                        & (0.18) & (0.14) & (0.22) & (0.32) & (0.45) &  \\
\addlinespace[0.2cm]
& \multirow{2}{*}{.7} & \multirow{2}{*}{72} & 0.73 & 0.45 & 0.90 & 1.25 & 1.52 &   \\
&  &                                        & (0.18) & (0.14) & (0.21) & (0.28) & (0.46) &  \\
\addlinespace[0.2cm]
& \multirow{2}{*}{.8} & \multirow{2}{*}{24} & 1.43 & 0.88 & 1.68 & 2.47 & 3.45 &  \\
&  &                                        & (0.44) & (0.29) & (0.50) & (0.86) & (1.62) &  \\
\addlinespace[0.2cm]
\multirow{2}{*}{\bfseries Subtraction}
& \multirow{2}{*}{.8} & \multirow{2}{*}{48} & 1.75 & 1.10 & 1.98 & 3.08 & 4.04 &  \\
&  &                                        & (0.42) & (0.26) & (0.45) & (0.93) & (1.52) &  \\
\addlinespace[0.2cm]
& \multirow{2}{*}{.8} & \multirow{2}{*}{72} & 1.85 & 1.16 & 2.12 & 3.29 & 4.20 &  \\
&  &                                        & (0.42) & (0.26) & (0.47) & (0.92) & (1.42) &   \\
\addlinespace[0.2cm]
& \multirow{2}{*}{.9} & \multirow{2}{*}{24} & 1.95 & 1.29 & 2.21 & 3.29 & 4.37 &  \\
&  &                                        & (0.36) & (0.32) & (0.48) & (0.67) & (1.19) &  \\
\addlinespace[0.2cm]
& \multirow{2}{*}{.9} & \multirow{2}{*}{48} & 2.11 & 1.42 & 2.34 & 3.59 & 4.51 &  \\
&  &                                        & (0.25) & (0.24) & (0.33) & (0.55) & (0.89) &  \\
\addlinespace[0.2cm]
& \multirow{2}{*}{.9} & \multirow{2}{*}{72} & 2.17 & 1.49 & 2.40 & 3.68 & 4.52 &  \\
&  &                                        & (0.24) & (0.20) & (0.31) & (0.55) & (0.91) &   \\
\addlinespace[0.2cm]
\bottomrule
\addlinespace[0.2cm]
\multicolumn{9}{r}{($n=300$ for each group)}
\end{tabular}
\end{table*}

\end{document}